\RequirePackage{fix-cm}
\documentclass[twocolumn]{svjour3}          
\usepackage{graphicx}
\usepackage{amssymb}
\usepackage{bm}
\usepackage{amsfonts}
\usepackage{amsmath}
\usepackage{amssymb}
\usepackage{natbib}
\usepackage{graphics}
\usepackage{fancyhdr} 
\usepackage{ucs}
\usepackage[utf8]{inputenc}
\usepackage{algorithmic}\usepackage[ruled,vlined]{algorithm2e}
\usepackage[colorlinks,hyperindex,bookmarks,linkcolor=blue,citecolor=blue,urlcolor=blue]{hyperref}
\usepackage{multicol, blindtext}
\usepackage{multirow}
\usepackage{booktabs}
\usepackage[justification=centering]{caption}
\usepackage[table]{xcolor}
\usepackage{latexsym}
\usepackage{color,soul}
\usepackage{float}
\usepackage{placeins}
\usepackage{caption}

\begin{document}

\title{End-to-end Inception-Unet based Generative Adversarial Networks for Snow and Rain Removals
}


\author{Ibrahim Kajo* \and
        Mohamed Kas* \and
        Yassine Ruichek    
}


\institute{Ibrahim Kajo, Mohamed Kas, and Yassine Ruichek \at
CIAD, UMR 7533, Univ. Bourgogne Franche Comté, UTBM, F 25200, Montbéliard, France. \email ibrahim.kajo@utbm.fr, \email mohamed.kas@utbm.fr, \email yassine.ruichek@utbm.fr. \and *Authors contributed equally.              
}

\date{Received: date / Accepted: date}

\maketitle
\begin{abstract}
The superior performance introduced by deep learning approaches in removing atmospheric particles such as snow and rain from a single image; favors their usage over classical ones. However, deep learning-based approaches still suffer from challenges related to the particle appearance characteristics such as size, type, and transparency. Furthermore, due to the unique characteristics of rain and snow particles, single network based deep learning approaches struggle in handling both degradation scenarios simultaneously.  In this paper, a global framework that consists of two Generative Adversarial Networks (GANs) is proposed where each handles the removal of each particle individually. The architectures of both desnowing and deraining GANs introduce the integration of a feature extraction phase with the classical U-net generator network which in turn enhances the removal performance in the presence of severe variations in size and appearance. Furthermore, a realistic dataset \footnote{\href{https://github.com/KasMohamed-tech/Inception-UNet-GAN-for-Snow-and-Rain-Removals}{
Inception-UNet-GAN-for-Snow-and-Rain-Removals}} that contains pairs of snowy images next to their groundtruth images estimated using a low-rank approximation approach; is presented. The experiments show that the proposed desnowing and deraining approaches achieve significant improvements in comparison to the state-of-the-art approaches when tested on both synthetic and realistic datasets.
\end{abstract}

\section{Introduction}
\label{intro}

The Performance of several computer vision applications such as object detection, object tracking, and semantic segmentation is highly dependent on the perceptual as well as the visual quality of the processed images. Therefore, bad weather conditions such as rain, fog, and snow have a negative impact on most image/video processing algorithms due to the image degradations caused by such conditions. Numerous learning-based approaches show better restoration performance in the presence of different weather degradations such as rain \citep{2} and snow \citep{3}, \citep{4} when compared to hand-crafted video based ones \citep{tian2018snowflake,santhaseelan2015}. Due to the difficulty of obtaining pairs of degraded images and their corresponding clean images, the majority of learning-based approaches including GANs are trained, tested and evaluated on datasets that consist of pairs of clean images and their synthetic degraded images. However, synthetic images fail to provide the level of authenticity that realistic images have. Hence, training networks on such synthetic datasets ends up in techniques that perform well on images degraded by synthetic rain masks, while fail drastically when they are applied on realistic images \citep{first}. Moreover, the availability of realistic test images allows more effective and accurate evaluation of existing deraining and desnowing approaches.  
For example, a recent study \citep{2} that reviewed most of the state-of-the-art deraining approaches shows that only two techniques (out of 27 reviewed) are trained and quantitatively evaluated using realistic datasets. To address this challenge, Wang et al. \citeyearpar{5} introduced a large-scale realistic rain/rain-free dataset for training and evaluating learning-based deraining techniques. Despite the availability of such an efficient dataset, a significant number of deraining approaches \citep{6,7,8,9,huang2021selective} have not considered this dataset (or other realistic datasets) for training better models or evaluating them quantitatively. On the other hand, the existing desnowing approaches lack the availability of a testing dataset that consists of realistic snow/snow-free pair images that contain a large variety of snowflakes in terms of size and appearance.

Most recently, different GANs based approaches have been proposed for single image desnowing and deraining where distinct architectures, descriptors, and loss functions are introduced \citep{10,11,12}. Generally, a single GANs framework consists of a generator network and a discriminator network, which mainly judges the image generated by the generator whether it is real or fake. However, the generator is also guided by several loss functions that provide meaningful feedbacks about the accuracy of its generated images. Multiscale pixel-level loss, structural similarity index (SSIM) loss, refined perceptual loss, and multiscale perceptual loss are good examples of such loss functions that are classically added to the objective function of a GANs framework. However, such loss functions are region blind, where snow/rain-free regions have the same impact as regions with snow/rain on minimizing the generation loss. In other words, the generator is not well guided to focus on the regions that are mainly occupied by snow/rain particles. Therefore, we manage in this paper to tackle this challenge by introducing a spatially guided loss that forces the generator to pay more attention to the snow/rain regions. This is mainly achieved by defining a loss that measures the similarity between the generated and the ground-truth snow/rain maps rather than their pixel-wise images. 

Due to the unique aspects of snowflakes compared to rain drops in terms of size, shape, and level of transparency, most learning-based approaches that are designed to remove snowflakes cannot simultaneously remove the rain streaks and vice versa. Therefore, developing one model that represents all image degradations caused by both rain and snow is a non-trivial task. Hence, we introduce a deraining/desnowing framework that consists of two degradation removers. 
The input images are fed into the corresponding network to remove the occurred degradation. In this paper, the proposed desnowing and deraining networks are designed based on the GANs architecture. However, each network has its unique architecture, generator and discriminator networks, and loss function that make them fundamentally different from other existing deraining and desnowing GANs. More specifically, the contributions of the paper are as follows:

\begin{itemize}
    \item  A novel architecture that introduces the usage of a feature extraction phase to the classical U-net generator network; is proposed. Such a modification enhances the image generation performance in handling a variety of sizes and appearances of snow and rain particles.
    
    \item A novel GAN architecture that allows the collaboration between two generators to derain and refine complex rainy images is introduced.
    
    \item Two novel loss functions that respect the unique characteristics of both snowflakes and rain particles; are designed to spatially guide the GAN generators to focus its removal performance toward corrupted regions. 
    
    \item A realistic test dataset for evaluating desnowing learning approaches is introduced where it is generated based on video-based low-rank approximation and human supervision.
\end{itemize}

\section{Related work}
\label{related_works}

\subsection{Desnowing techniques }

Due to their efficient generalization abilities, learning based techniques have been recently employed to remove snowflakes, rain streaks, and rain drops. However, researchers tend to propose unique removal networks that consider the unique characteristics of each particle separately. Liu et al \citeyearpar{3} proposed the first learning based technique that is dedicated for removing snowflakes from a degraded image. They designed a multistage network named DesnowNet to deal with images that are degraded by both translucent and opaque snowflakes. The first stage of their network considers recovering the translucent snowflakes using a translucency recovery (TR) module while the second stage considers recovering the opaque snowflakes employing the residual generation (RG) module. In order to train their network, they proposed the first dataset that contains enough synthetic snowy images along with their snow masks and ground-truths in addition to a realistic testing dataset with no ground truth provided. Chen et al. \citep{4} proposed a single image desnowing model that considers several appearance characteristics such as size, transparency, and veiling effect. Their framework starts with an identifier that consists of three end-to-end networks to detect and classify the snowflakes into small, medium, and large particles. Additionally, a dark channel based prior layer is embedded into an end-to-end network to help deal with the veiling effect. The detected initial snow masks and the veiling effect-free images are further processed by a transparency-aware snow removal model that is inspired by an inpainting mechanism to recover the original pixels occluded by non-transparent snowflakes. To optimize the results, a size-aware discriminator is designed to discriminate between snowy and snow-free images. Jaw et al \citeyearpar{13} introduced a dual pathway descriptor that extracts the semantic features via  a bottom-up pyramidal pathway and aggregates the features extracted at different resolutions to provide more accurate location information that enhances the snow removal process and reduces the computational burden. Their snow removal network is followed by a GAN based refinement stage to provide more detailed images. 

However, all aforementioned learning approaches are trained and quantitatively evaluated using only synthetic datasets since real ground-truth images are not available. To fill this gap in this research area, we introduced a realistic dataset that consists of real snow/snow-free images estimated by a video-based low-rank approximation approach. This dataset can be combined with other synthesis snowy images to enrich the learning process during training and provides a chance to test and quantitatively evaluate the proposed desnowing approaches including the proposed one.

\subsection{Deraining techniques}

There are several deraining learning techniques which have been proposed recently in the literature \citep{14,15,16,17,18}.  Yang et al. \citeyearpar{19} introduced a single image deraining framework in which the rain streak regions are detected and added as a binary map to the developed model. The detected rain streaks are accumulated to simulate the atmospheric veil so the designed framework can generate more accurate visual results. In their network, a contextualized dilated network is developed to extract the discriminative rain features and enhance the stages of rain detection and removal. To handle real-world rain images and visually enhance the background layer, a cascade of several convolutional joint networks of rain detection and removal is recurrently employed. Fu et al. \citeyearpar{20} proposed a single image deraining deep detail network with an architecture inspired by the deep residual network (ResNet). They introduced a lossless negative residual mapping to replace the traditional loss function used in ResNet model. Their negative mapping is computed based on the difference between the rainy image and the rain-free image reducing the range of the pixel values which in turn speeds up the deraining learning process. The input image is converted into a detailed layer using a high-frequency filter which in turn is fed to the parameter layer to predict the rain mask. 

However, the majority of the deraining learning based techniques were trained on synthesized datasets with limited realism in respect to rain characteristics such as appearance, direction and transparency. To address the problem related to the availability of realistic training paired images, Lin et al. \citeyearpar{21} introduced a deraining approach inspired by data distillation principles. The proposed approach does not require paired images for training where unpaired set of rainy and clean images is utilized. The clean images are synthesized with the same rain masks that degrade the input images. To achieve this purpose, blurred rain-free images are fed into an enhancement learning block that extracts both the rain and the detail layers. The extracted rain layers are added to the clean images, and the degraded images are fed back into a deraining network to correctly generate the rain mask of the input images. Following a different approach, Wang et al. \citeyearpar{5} utilized a semi-automatic filtering method that incorporates both temporal priors and human supervision to generate a large training dataset with accurate ground-truth images at high resolution and free of rain degradation. Furthermore, they proposed a spatial attentive rain removal network guided by recurrent neural networks with RelU and identity matrix initialization (IRNN). The introduced dataset provides a helpful tool to train robust and efficient learning models that can be implemented in real-world scenarios. However, their network introduces an expensive computational complexity in terms of memory where a large number of learnable parameters is used.

The  success  of GANs based frameworks in generating visual appealing results  has  inspired  researchers  to  propose and explore dozens of GANs’ architectures for image synthesize purposes including desnowing \citep{12,22}. Zhang et al. \citeyearpar{12} employed a GAN guided by a refined perceptual loss to enhance the structure awareness ability of the refinement performance of their proposed deraining approach. Cai et al.\citeyearpar{10} introduced a depth-density based GAN guided by the depth and density information provided by another network in order to achieve better rain streak and fog removal performance. Ding et al \citeyearpar{11} proposed a GAN that accepts 3D light field images (LFI) as input where the disparity maps are first estimated and the rain streaks are extracted via two branched autoencoders. This approach successfully takes full advantage of the structural information embedded in LFIs to remove the rain streaks. Li et al. \citeyearpar{23} proposed a method that can handle rainy, snowy, and foggy images using a single network with three dedicated encoding streams that process the same input image. Unlike typical adversarial learning, the introduced discriminator architecture is utilized to identify the degradation type in addition to the status of the image (clean or noisy). However, such a solution employs more sophisticated decoding architectures and feature search strategies in order to take all the encoding streams of multiple degradations into account, imposing more computational burdens in terms of memory and time.

The generators of all these deraining GANs are guided by region-blind loss functions, where the majority of these losses are designed to measure the similarity between the generated image and its groundtruth at pixel levels . In addition to this, our proposed deraining GAN consists of two generators where the first one is spatially guided to focus on recovering the background pixels of the regions occupied by the rain particles, while the second generator's task is to visually enhance the result estimated by the first generator.   

\section{Proposed autoencoder of a generator network}

\subsection{Brief overview of GANs}

As mentioned earlier in this paper, we propose two conditional GAN-based networks to retrieve the original clean image $C$ from a degraded image $X$ in the case of rain and snow scenarios. Both the desnowing and deraining frameworks consist of a GAN-based architecture. The generative adversarial networks have been proposed in the first time by Godefellow \citeyearpar{24} as a tool to generate new images that look like images belonging to the target space. The networks developed based on the GAN principles are designed to model certain data distributions. To achieve this purpose, a generator network $\mathbf{G}$ is employed to regenerate samples that share the same data distribution of the target data. Additionally, a discriminator network $\mathbf{D}$ is incorporated to measure the probability of the generated samples having the same data distribution or not. Based on the game theoretic $min-max$ principle, the generator and discriminator are typically learned jointly by alternating the training of $\mathbf{D}$ and $\mathbf{G}$.

\subsection{End-to-end Inception-Unet based generator}

The generator network $\mathbf{G}$ relies on autoencoder architecture, where U-Net \citep{25} and ResNet-Blocks \citep{26} architectures are widely used by the majority of the state-of-the-art GANs. The concept of U-Net architecture is to supplement a downsampling stream by symmetric layers (decoding) where traditional pooling operations are replaced by transposed convolution-based upsampling giving the final architecture the shape of letter U. As a result, these layers increase the resolution of the decoded output with more precision thanks to the skip connections. Moreover, U-Net provides a large number of feature channels in the upsampling part, which allows the network to propagate context information to higher resolution layers. All the U-Net variants share the same pipeline, the only difference is the supported image size which is controlled by the amount of GPU memory where most of the frameworks consider the resolutions of 256 and 128.

On the other hand, ResNet-based generator networks adopt residual blocks to compute relevant features from the input images. Hence, the input image is downsampled generally by 2 or 3 before the resulted convoluted feature maps are fed into residual sub-networks referred to as ResNet-Blocks. Afterwards, the output of a sequence of ResNet-Blocks is upsampled to reach the desired size of the generator output. However, both architectures have their limitations in terms of computational complexity and implementation. For example, the U-Net generator suffers from several converging difficulties during training due to the lack of deeper feature extraction blocks. On the other hand, ResNet based generators suffer from a major drawback where the resultant images are blurred despite the presence of their feature extraction sub-blocks. Moreover, ResNet generators have no skip connections which are very beneficial especially when the target and input images share some visual features, as the case in our task.

\begin{figure*}[!ht]
    \centering
    \includegraphics[width=17cm, height=9.5cm]{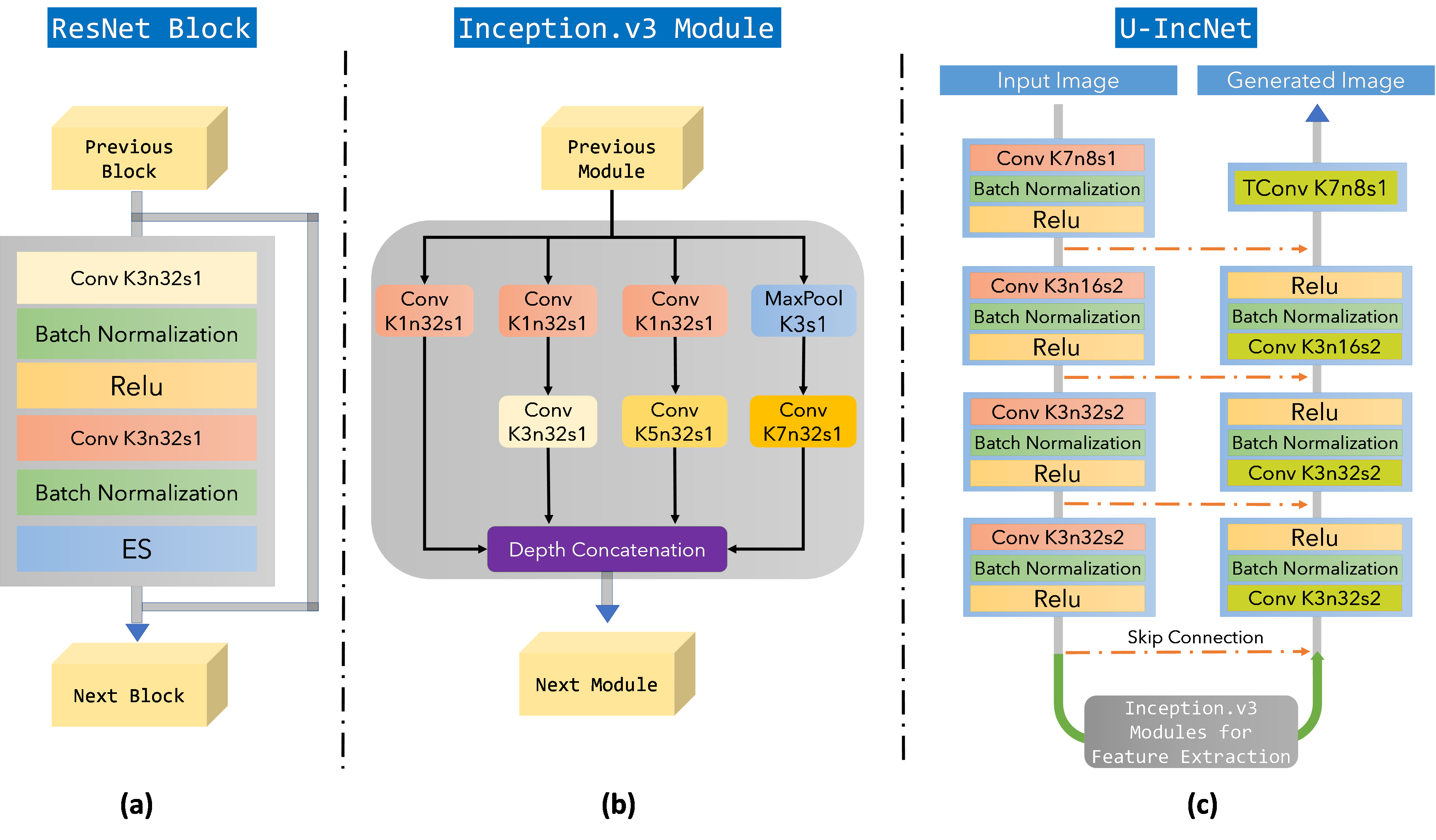}
    \caption{Visual architecture-based comparison of: classical ResNet feature extraction mechanism (a), Inception.v3 based feature extraction (b), and the proposed U-IncNet (c).}
    \label{Fig1}
\end{figure*}

To tackle these challenges, we introduced a U-net inspired architecture, called U-shaped inception-based network (\textbf{U-IncNet}), that includes additional feature extraction blocks which link the downsampling and upsampling streams. The feature extraction sub-blocks involve several \textbf{Inception.v3} \citep{27} modules. In contrast to Resnet block, Inception.v3 module computes the convolution responses with multi-kernel sizes: $\{(1 \times 1), (3 \times 3), (5 \times 5), (7 \times 7) \}$. These filters are concatenated to form the final output of the Inception.v3 modules. The choice of Inception.v3 module as feature extractor is based on its capabilities of capturing visual features at different scales thanks to the different kernel sizes. Incorporation of Inception.v3 proves its significance in the field of deraining/desnowing applications, where large kernels $(5 \times 5)$ and $(7 \times 7)$ help the model remove the large rain/snow patterns that may appear like parts of the background. Fig \ref{Fig1} shows the difference between the classical ResNet feature extraction mechanism (see Fig. \ref{Fig1} (a)) and Inception.v3 based feature extraction (see Fig. \ref{Fig1} (b)) in addition to the proposed U-IncNet (see Fig. \ref{Fig1} (c)). Finally, the discriminator network $\mathbf{D}$ relies on a pixel classification architecture. It is based on a set of convolutions and activation layers that lead to a binary label (fake/real) for each pixel. The state-of-the-art discriminator networks demonstrated good performances in detecting fake and real images, and then granting prominent adversarial loss to the generator. In our work, we use the Pixel Discriminator architecture \citep{28}.

\section{Weather-degradation removal networks}
The appearance of a degraded image X due to rain or snow conditions is mathematically modeled as a summation of two basic layers where the first layer represents the original background, while the second layer represents the rain/snow layer. Additional layers are introduced to the appearance model to represent several accompanied atmospheric conditions such as fog, veiling effect, reflection, occlusion, and rain accumulation. To simplify the representation of the degradation problem, we assume that a degraded image $X$ can be defined as follows:
\begin{equation}
\label{eq1}
    X=C+N  
\end{equation}
where $C$ refers to the degradation-free background image and $N$ refers to the summation of all degradation layers including rain/snow and other accompanied atmospheric conditions. The objective of the proposed desnowing or deraining network is to estimate the clear background image $C$ by removing the degrading layers $N$ from the input image $X$.
Our approach of snow/rain removal is based on handling each condition individually. As highlighted earlier, splitting the conditions (rain/snow) allows efficient image enhancement as well as low computational cost. 
Hence, X is fed to the appropriate GAN to eliminate the degradations. This subsection is divided into 2 subsections to explain in depth the architectures of the proposed Desnowing GAN and Deraining GAN.

\begin{figure}[!ht]
    \centering
    \includegraphics[width=8.3cm, height=8cm]{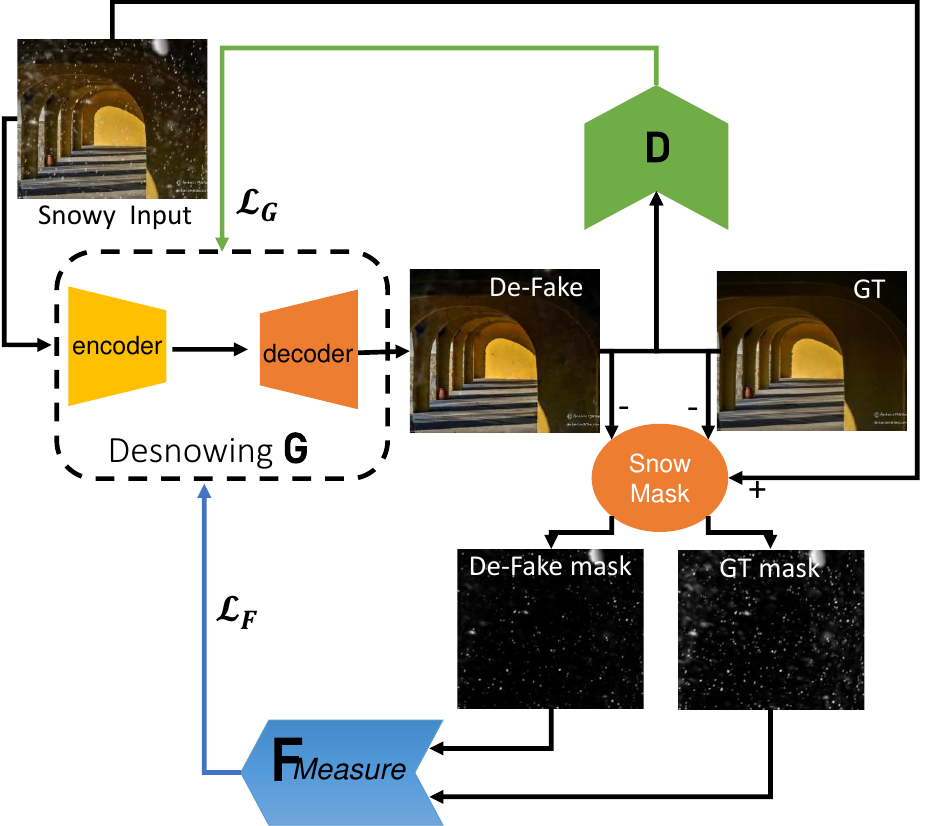}
    \caption{The proposed SGAN learning architecture}
    \label{F1loss}
\end{figure}

\subsection{Proposed desnowing Network (SGAN)}

The SGAN is a combination of two convolutional neural networks (CNN) where the first one is an image generator and the second is a pixel discriminator. Considering the snowy image $X_s$ as condition, the image generator $\mathbf{G}$ attempts to generate an image $\hat{C}=\mathbf{G}\left(X_s\right)$ similar to its corresponding snow-free image $C$ by encoding the input image $X_s$ into a latent space with discriminant features. Then, the encoded image is upscaled recurrently until producing the desired snow-free image $C$. On the other hand, the discriminator $\mathbf{D}$ that is a pixel binary classification network, learns to identify both the snow-free image $C$ and the fake one $\hat{C}$ produced by the generator $\mathbf{G}$. The discriminator output is fed back into the generator forcing it to generate a better image which should be similar to the snow-free image C. In other words, the discriminator $\mathbf{D}$ is supervising the generator $\mathbf{G}$, which requires the optimization of the discriminator at each epoch so it can judge whether the generated image $\hat{C}$ is fake or real. 

Similar to any CNN architecture, the training of GAN is based on optimizing its objective loss function $\mathbf{L}_{G}$, which is illustrated as follows 
\begin{equation}
\label{eq2}
  \mathbf{L}_{G}=arg\ min_\mathbf{G}{max_\mathbf{D}{(\log{\mathbf{D}(}}}C))+ \log{(1-\mathbf{D}(\hat{C}))}+ \lambda.\mathbf{L}_\mathbf{1}  
\end{equation}
where $\mathbf{L}_\mathbf{1}=\left\|C-\hat{C}\right\|_1$ represents the loss between the ground truth image and the generated one and $\lambda$ is a regularization parameter. $\mathbf{L}_\mathbf{1}$ loss is included into the objective function $\mathbf{L}_{G}$ to help the generator producing less blurred images. However, extracting a snow-free image from a snowy one faces the problem of the occlusion caused by opaque snowflakes which totally occlude the background pixels. To tackle this problem, the generator is spatially guided to generate an image which produces the same snow mask as the GT image when both images are subtracted from the same corresponding snowy image. To achieve this, we compute the similarity between the original residual R calculated by subtracting the snow-free image C from the snowy image $X_s$ and the estimated residual $\hat{R}$ calculated by subtracting the generated snow-free image $\hat{C}$ from the snowy image $X_s$. Subsequently, both the original residual R and the estimated residual $\hat{R}$ are binarized using thresholding to generate their corresponding snow masks. To measure the similarity between these masks, we propose incorporating $F_{Measure}$ due to its sensitivity to the position and shape of pixels. The numbers of true positives, false positives, and false negatives are calculated to compute both the precision and the recall which are needed to calculate the $F_{Measure}$ as follows:

\begin{equation}
\label{eq3}
    R=\frac{TP}{TP+FP}
\end{equation}
\begin{equation}
\label{eq4}
      P=\frac{TP}{TP+FN} 
\end{equation}
\begin{equation}
    \label{eq5}
    F_{Measure}=\frac{2\times R\times P}{R+P}
\end{equation}

The estimated $F_{Measure}$ values range between 0 and 1 where low values indicate low similarity and high values indicate high similarity between the compared snow masks. Based on that the $F_{Measure}$ loss can be defined as follows:

\begin{equation}
    \label{eq6}
    \mathbf{L}_F=\lambda_f.\ F_{Measure}
\end{equation}
where $\lambda_f$ is a predefined weight for the $F_{Measure}$ loss. Hence, the overall objective function $\mathbf{L}_{s}$ is defined as follows:
\begin{equation}
    \label{eq7}
\mathbf{L}_{s}=\mathbf{L}_{G}+\mathbf{L}_F
\end{equation}
For more clarity, Fig. \ref{F1loss} demonstrates the architecture of the proposed SGAN including the process of computing the proposed loss function.

\subsection{Proposed deraining Network (RGAN)}

\begin{figure*}[!ht]
    \centering
    \includegraphics[width=16cm, height=5cm]{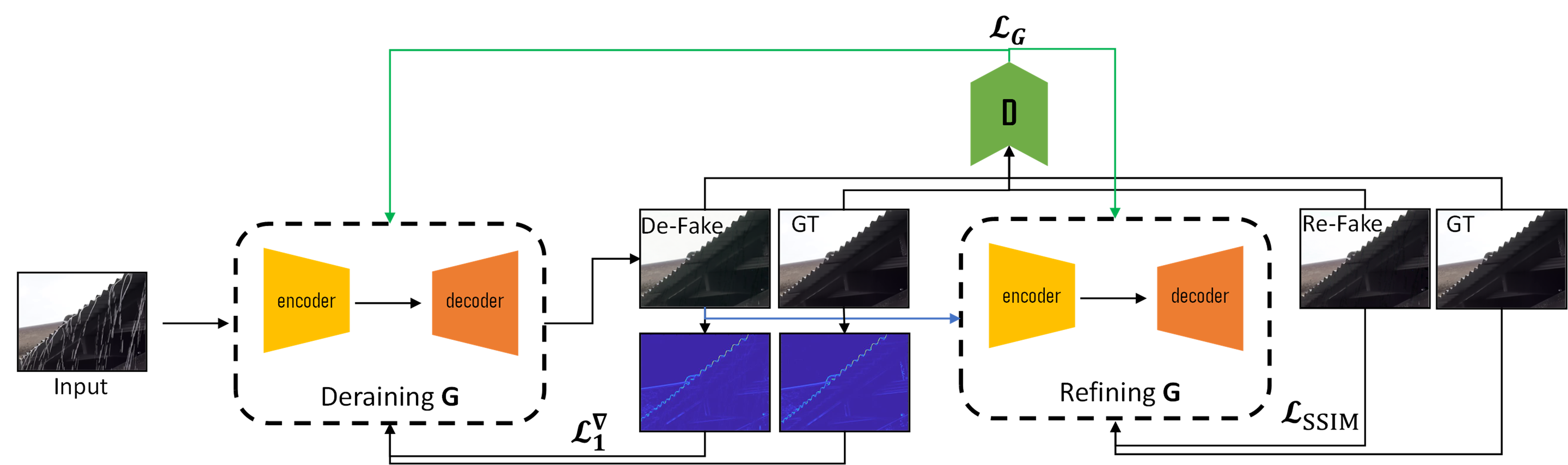}
    \caption{Overview of our proposed collaboration-based pipeline that consists of two generators where the first one $\mathbf{G}_{d}$ removes rain strikes/drops from input images while the second generator $\mathbf{G}_{r}$ corrects the visual artifacts that may be caused by the first generator.}
    \label{Fig2}
\end{figure*}

In contrast to snowy images, rainy images present more challenges that need to be handled simultaneously. A wide variety of rain pattern shapes (drops, streaks, accumulation), a variety of transparency levels, multiple falling orientations, in addition to the accompanied veiling effects; are obvious examples of the added complexity when solving the deraining problem. The majority of the deep learning approaches are trained on synthetic datasets which show several limitations in modelling the rain physically. Therefore, such learning approaches fail to effectively derain degraded images and produce unsatisfactory images in terms of visual and numerical results. 
Conducted preliminary experiments show that traditional GAN frameworks that are trained on both synthesis and realistic datasets continue to produce the same satisfactory results, especially when they are tested on realistic images. 

Taking all raised points into consideration, our RGAN is designed with more sophisticated architecture for better representation of the deraining problem. As depicted in Fig. \ref{Fig2}., the proposed RGAN is developed to contain two image generators instead of one where the input image of the first generator $\mathbf{G}_{d}$ is the rainy image $X_r$ and its output image $\acute{C}=\mathbf{G}_{d}\left(X_r\right)$ is fed as an input into the second generator $\mathbf{G}_{r}$ to generate the final derained image $\hat{C}$. Applying the appropriate objective function of each generator allows controlling the appearance of the image that each generator should generate. Hence, the first generator $\mathbf{G}_{d}$ is given the task of removing the rain patterns from an input image while the second generator ${\mathbf{G}_{r}}$ is given the task of correcting any color variation or artifacts generated by $\mathbf{G}_{d}$. Both generators share the same discriminator; therefore, they show a collaborative behavior where both generators attempt to fool the discriminator by generating more plausible images.

The objective function of the first generator $\mathbf{L}_{\mathbf{G}_{d}}$ contains the same $\mathbf{L}_{G}$ as in Eq (2) in addition to a new loss function $\mathbf{L}_\mathbf{1}^\mathbf{\nabla}$ introduced to guide the generator to generate better deraining images. Due to the unique shape and appearance characteristics of the rain patterns, $\mathbf{L}_\mathbf{1}^\mathbf{\nabla}$ is designed based on the gradient magnitude maps of both the rain-free image $C$ and the estimated rain-free image $\acute{C}$ generated by ${\mathbf{G}_{d}}$, which are computed as follows:

\begin{equation}
    \label{eq8}
    \left \| \nabla C \right\| = \sqrt{(\frac{\partial C}{\partial x})^{ ^2} + \alpha \times (\frac{\partial C}{\partial y})^{ ^2}}
\end{equation}

\begin{equation}
    \label{eq9}
    \left \| \nabla \acute{C} \right\| = \sqrt{(\frac{\partial \acute{C}}{\partial x})^{ ^2} + \alpha \times  (\frac{\partial \acute{C}}{\partial y})^{ ^2}}
\end{equation}
where ${\alpha}$ is a weighting parameter that gives more impact on the vertical gradient to better represent the rain streaks. Hence, $\mathbf{L}_\mathbf{1}^\mathbf{\nabla}$ is the $\mathbf{L}_\mathbf{1}$ loss between the gradient magnitude map of the ground truth and the gradient magnitude map of the estimated image as follows: 
\begin{equation}
    \label{eq10} 
    \mathbf{L}_\mathbf{1}^\mathbf{\nabla}=\lambda_{G_d} \times \left \| {\left \| \nabla C \right\| - ||\nabla \acute{C}||} \right\|
\end{equation}
where ${\lambda_{G_d}}$ is a predefined weight. Hence, the cost function $\mathbf{L}_\mathbf{1}^\mathbf{\nabla}$ is employed in the objective function of the generator $\mathbf{G}_{d}$ as follows: 
\begin{equation}
    \label{eq11} \mathbf{L}_{\mathbf{G}_{d}}=\mathbf{L}_{G}+\mathbf{L}_\mathbf{1}^\mathbf{\nabla}
\end{equation}

Similarly, the objective function $\mathbf{L}_{\mathbf{G}_{r}}$ of the generator $\mathbf{G}_{r}$ consists of the basic loss $\mathbf{L}_G$ in addition to structural similarity index measure (SSIM) based loss which is defined as follows:

\begin{equation}
    \label{eq12}
    \mathbf{L}_{SSIM}=\frac{(2\mu_C\mu_{\hat{C}}+c_1)(2\sigma_{C\hat{C}}+c_2)}{(\mu_C^2+\mu_{\hat{C}}^2+c_1)(\sigma_C^2\sigma_{\hat{C}}^2+c_2)}
\end{equation}
where $C$ and $\hat{C}$ represent the rain-free image and the estimated enhanced image generated by ${\mathbf{G}_{r}}$, respectively. $\mu$ and $\sigma$ refer to the mean and the variance of the corresponding image, respectively.  $c_1$ and $c_2$ are constants. Thus, the objective function $\mathbf{L}_{\mathbf{G}_{r}}$ can be defined as follows:
\begin{equation}
    \label{eq13}
    \mathbf{L}_{\mathbf{G}_{r}}=\mathbf{L}_G+\mathbf{L}_{SSIM}
\end{equation}

\section{SVD-based ground truth generation of realistic snow images}
\label{sect4}

Providing snow-free/snowy pair images to train learning models is a challenging task for most of the proposed approaches. Capturing the same scene before/after the snow falling event provides images which are free of both falling and static snowflakes. However, changes in lighting and appearance may occur and take place in the extracted pairs, leading to poor learning process of the trained models. In contrast to deraining scenarios, several testing experiments of the existing state-of-the-art desnowing approaches (including our proposed one) have shown that training the networks on synthetic datasets allows the trained networks effectively removing the snow layer. However, realistic snow-free images are still needed for the purpose of more comprehensive evaluations and comparisons. Therefore, we propose a ground truth image extraction scheme from snowy videos captured by a static camera. To simplify such extraction and avoid unnecessary foreground removal, the selected videos contain only snowflakes as moving objects. We have downloaded 56 high resolution $1920 \times 1080$ videos from the internet and split each video image into four $960 \times 540$ sub-images to increase the number of test images. The extracted sub-images are further processed by the proposed ground-truth extraction scheme.


\begin{figure*}[!ht]
    \centering
    \includegraphics[width=12cm, height=6cm]{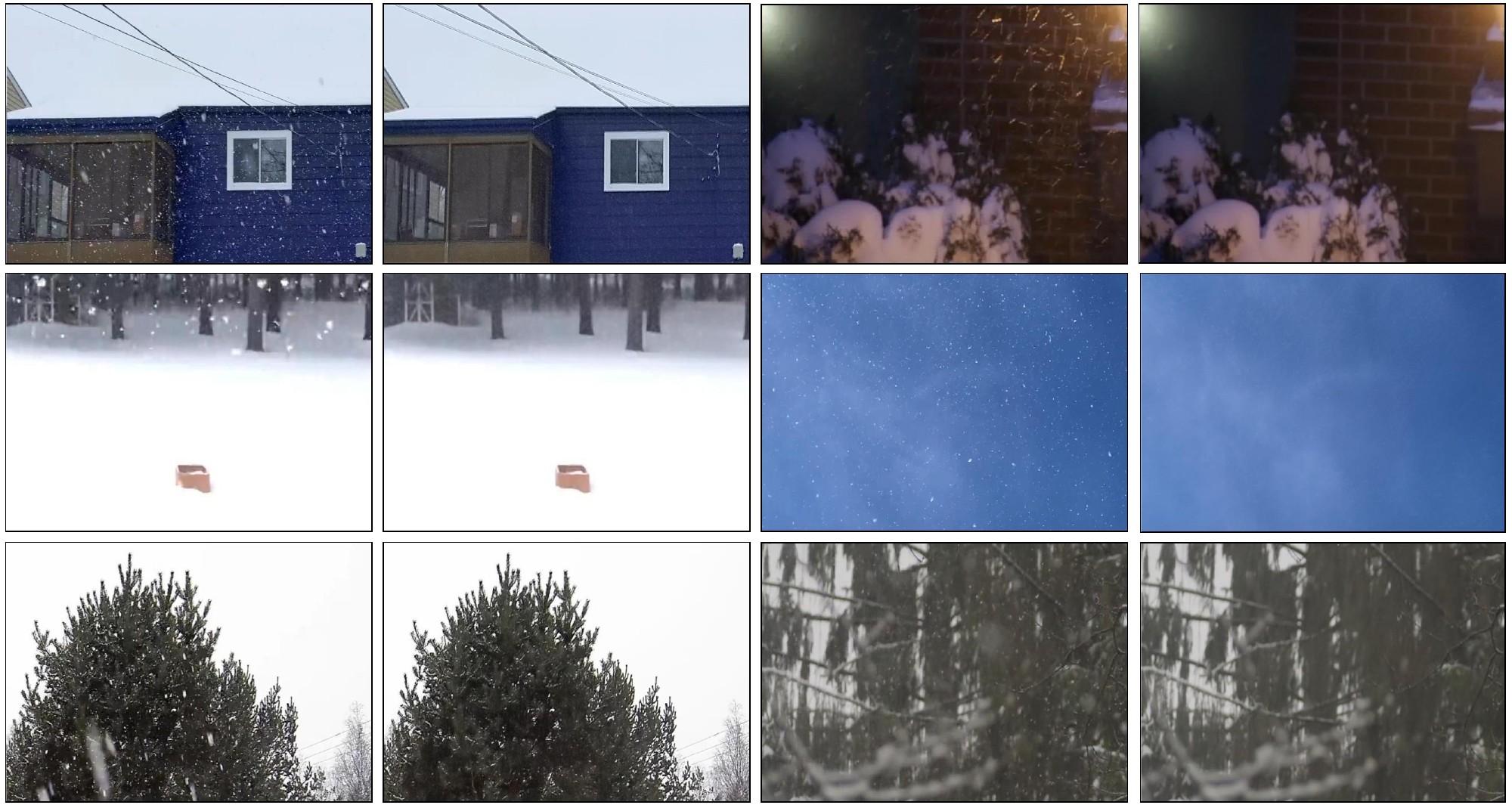}
        \caption{Visual examples of the realistic dataset: $1^{st}$ and $3^{rd}$ columns show snowy images while $2^{nd}$ and $4^{th}$ show their corresponding estimated ground-truths.}
    \label{realistic-dataset}
\end{figure*}

\begin{figure*}[!ht]
    \centering
    \includegraphics[width=12cm, height=6cm]{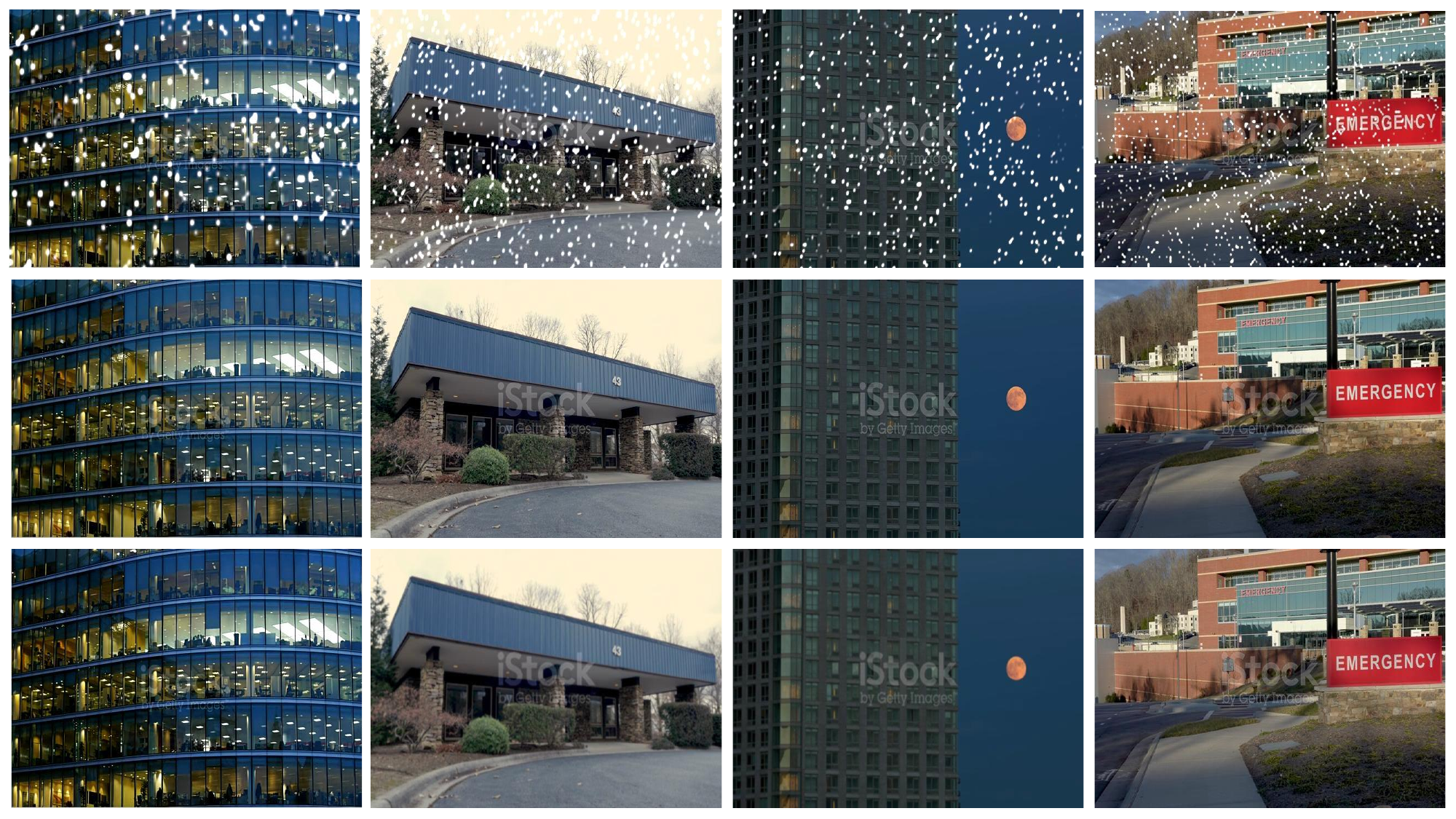}
        \caption{Visual examples of the snow removal results extracted from synthetic snowy videos: $1^{st}$ row shows several snowy frames, $2^{nd}$ row shows the original ground-truth, and $3^{rd}$ row shows the estimated ground-truths using our video based snow removal}
    \label{synthetic-dataset}
\end{figure*}

\textbf{Singular value decomposition (SVD)} \citep{29} is a powerful mathematical tool for statistically decomposing matrices into their singular components. SVD is widely used in the matrix/tensor approximation where data can be separated into different low-rank and sparse components. To remove the falling snow in a given video, we designed a snow removal SVD-based method that can remove the falling snowflakes and maintain the background. The proposed method depends on the fact that snowflake patterns behave differently than the background patterns with respect to both space and time dimensions. Therefore, the spatiotemporal slices of a given video tensor are decomposed by the proposed method into a low-rank component that represents the background information and a sparse component that represents the snowflakes. 

Consider the video tensor $\mathrm{\mathcal{A}\ \in\ }\mathbb{R}^{m\times n\times k}$, where $m$ indicates the number of rows, $n$ indicates the number of columns, and $k$ indicates the number of frames. Based on the selected dimension, three slice modes can be extracted: horizontal slices $\mathrm{\mathcal{A}}_{\ i::} \in R^{k\times n}$, lateral $\mathrm{\mathcal{A}}_{:j:}\in R^{k\times m}$ , and frontal $ \mathrm{\mathcal{A}}_{::k}\in R^{m\times n}$. Taking into account the spatiotemporal nature of both modes $\mathrm{\mathcal{A}}_{:j:}$ and $\mathrm{\mathcal{A}}_{::k}$, decomposing one of them into low-rank and sparse components allows separating the background layer from the snowflake layer. Based on this, the SVD of a spatiotemporal slice $\mathrm{\mathcal{A}}_{i::}$ can be expressed as follows:
\begin{equation}
    \label{eq14}
    \mathrm{\mathcal{A}}_{i::}=\mathbf{US}\mathbf{V}^T
\end{equation}
where matrices $\mathbf{U}\in R^{k\times k}$ and $\mathbf{V}\in R^{n\times n}$ have orthogonal columns known as the left and right singular vectors of $\mathrm{\mathcal{A}}_{i::}$, respectively. And matrix $\mathbf{S}\in R^{k\times n}$, is diagonal with real non-negative diagonal entries $\sigma_1\geq\sigma_2\geq\ldots\geq\sigma_p\geq0$ where $p=min\left\{k,n\right\}$. Hence, the spatiotemporal slice $\mathrm{\mathcal{A}}_{i::}$ can be reconstructed as a summation of $d$ projections  as follows:
\begin{equation}
    \label{eq15}
    \mathrm{\mathcal{A}}_{i::}={u}_1\sigma_1{{v}_1}^T+\ldots+{u}_d\sigma_d{{v}_d}^T
\end{equation}
where $d<p$ is the number of nonzero singular values. 
Based on the comprehensive analysis reported in \citep{30}, the first rank-1 projection $\mathrm{\mathcal{B}}_{i::} = u_1s_1v_1^T$ represents the content of stationary pixels throughout the time dimension of a video (i.e. the background) while the sum $\mathrm{\mathcal{F}}_{i::}=\sum_{l=2}^{q}{{u}_ls_l{v}_l^T}$ represents the sparse information that denotes the foreground objects. The residual, that is, the sum $\mathrm{\mathcal{N}}_{i::}=\sum_{l=q+1}^d u_ls_lv_l^T$, represents additive white Gaussian noise, such as illumination changes and thermal and quantization noise. Thus, a proper temporal filtering applied on the foreground slice should remove the snowflakes represented by short-term motion patterns. This is achieved by temporally filtering the left singular vectors that represent the motion information as follows: 
\begin{equation}
    \label{eq16}
    {\grave{{u}}}_l=f\left({u}_l\right)\ \ \ \ \ for\ l=2,3,\ldots,q
\end{equation}
where f is the filtering operator, which here is taken as an ideal temporal bandpass filter with sharp low- and high-cutoff frequencies denoted by $f_l$ and $f_h$, respectively. Consequently, the foreground components can be reconstructed using the filtered left singular vectors as follows: 
\begin{equation}
    \label{eq17}
    \mathrm{\mathcal{F}}_{i::}^{{desnowed}}\mathrm=\sum_{l=2}^q {\grave{{u}}}_l s_l {v}_l^T
\end{equation}
After applying the previous procedure on all horizontal slices, a video that contains no falling snowflakes can be constructed as follows:

\begin{equation}
    \label{eq18}
    \mathrm{\mathcal{A}}^{GT}=\mathrm{\mathcal{B}} + \mathrm{\mathcal{F}}^{{ desnowed}}  + \mathrm{\mathcal{N}}
\end{equation}

Then, we rely upon human judgment to select the best snow-free image that represents the ground truth of a given snow image. Figure \ref{realistic-dataset} shows several visual examples of the estimated ground-truth images extracted from snow images showing different types of snowflakes in terms of size and appearance. Furthermore, we implemented our video based desnowing on 10 synthesized snow videos  (snow video masks were generated using the tool provided in \cite{sadeghzadeh2021efficient}), where scores of 44.24/0.9914 of PSNR/SSIM are obtained by comparing the estimated results to the ground-truth background images. Figure \ref{synthetic-dataset} clearly demonstrates the effectiveness in extracting snow-free images from several synthesized videos.

\textbf{Dataset description}. A test subset that shows diverse background scenes (forests, buildings, and streets) with different snow densities are constructed. Using the aforementioned groundtruth generation scheme, we generate 83 pairs of snowy images of size $960 \times 540$ along with their corresponding groundtruth images.

\section{Experimental results}

The experiment section consists of two main experiments: the desnowing framework evaluation and the deraining framework evaluation. Each experiment contains ablation study and both numerical and visual comparisons to several state-of-the-art approaches using both synthesis and realistic datasets. For quantitative results, two basic evaluation metrics are used: peak-to-peak signal to noise ratio (PSNR) and structural similarity (SSIM). Both proposed desnowing and deraining approaches are implemented through Pytorch environment with GPU support. During the training, the batch size is set to 6 with 2000 iterations and the learning rate is fixed at 0.0001.

\subsection{Desnowing experiment}

\subsubsection{Ablation study}
The effectiveness of the SGAN components is demonstrated by conducting an ablation study in which different combinations of components are tested, as shown in Table \ref{ablationsnow}. In this ablation study, the proposed desnowing framework is trained and tested using the publicly available dataset Snow100K2 which contains 100k synthesized snowy images using 5.8k snow masks. Both the training and the testing sets are divided based on the 
size of the snowflakes into three subsets: Snow 100k-S (small snowflakes), Snow 100k-M (small and medium snowflakes), Snow 100k-L (small, medium, and large snowflakes). Tested variants are trained on a set that combines all three training subsets and tested on each subset individually.
First, the backbone autoencoder architecture (Unet) is trained and tested. In the second experiment, the base architecture is reinforced by the inception module (U-IncNet) using the traditional loss function. Thanks to the introduced feature extraction module, the desnowing performance is further enhanced compared to the results obtained by the baseline. The last experiment concerns studying the impact of the proposed loss $\mathbf{L}_F$ where better desnowing performance is expected due to the spatial guidance introduced during the learning process. The estimated results show the improvement provided by the incorporation of such loss where higher PSNR/SSIM values are obtained.

\subsubsection{Comparison with state-of-the-art approaches}

\textbf{Synthetic dataset}. Furthermore, our U-IncNet based desnowing framework is compared with state-of-the-art approaches and the quantitative results are listed in Table \ref{Tab2}. 
\begin{table}[htbp]
  \centering
  \caption{Evaluation of the effectiveness of the proposed U-IncNet network and the proposed $\mathbf{L}_F$ }
    \begin{tabular}{ll}
    \toprule
    \multicolumn{1}{c}{\textbf{method}} & \multicolumn{1}{c}{\textbf{PSNR/SSIM}} \\
    \midrule
    baseline (\#Params 54M) & 29.98/0.9372 \\
    baseline+U-IncNet (\#Params 32M) & 30.35/0.9371 \\
    baseline+U-IncNet+$\mathbf{L}_F$ & \textbf{31.93}/\textbf{0.965} \\
    \bottomrule
    \end{tabular}%
  \label{ablationsnow}%
\end{table}%
In this experiment, we compare our results with the results reported in \citep{3,13}  where several learning approaches which were originally proposed to handle problems such as deraining, dehazing, and semantic segmentation; are evaluated. However, the aforementioned approaches were retrained on the Snow 100k train set and their desnowing performances are evaluated. The quantitative results show that approaches such as DehazeNet \citep{31}, DerainNet \citep{32}, JORDER \citep{19} and DeepLab \citep{33} demonstrate poor performance in recovering clean images from snowy ones while DuRN-S-P \citep{34} approach shows satisfactory results in terms of PSNR and SSIM values. The recently proposed desnowing approaches DeSnowNet \citep{3} and DS-GAN \citep{13} show good performance in recovering snow-free images.

\begin{figure*}[!ht]
    \centering
    \includegraphics[width=17cm, height=8cm]{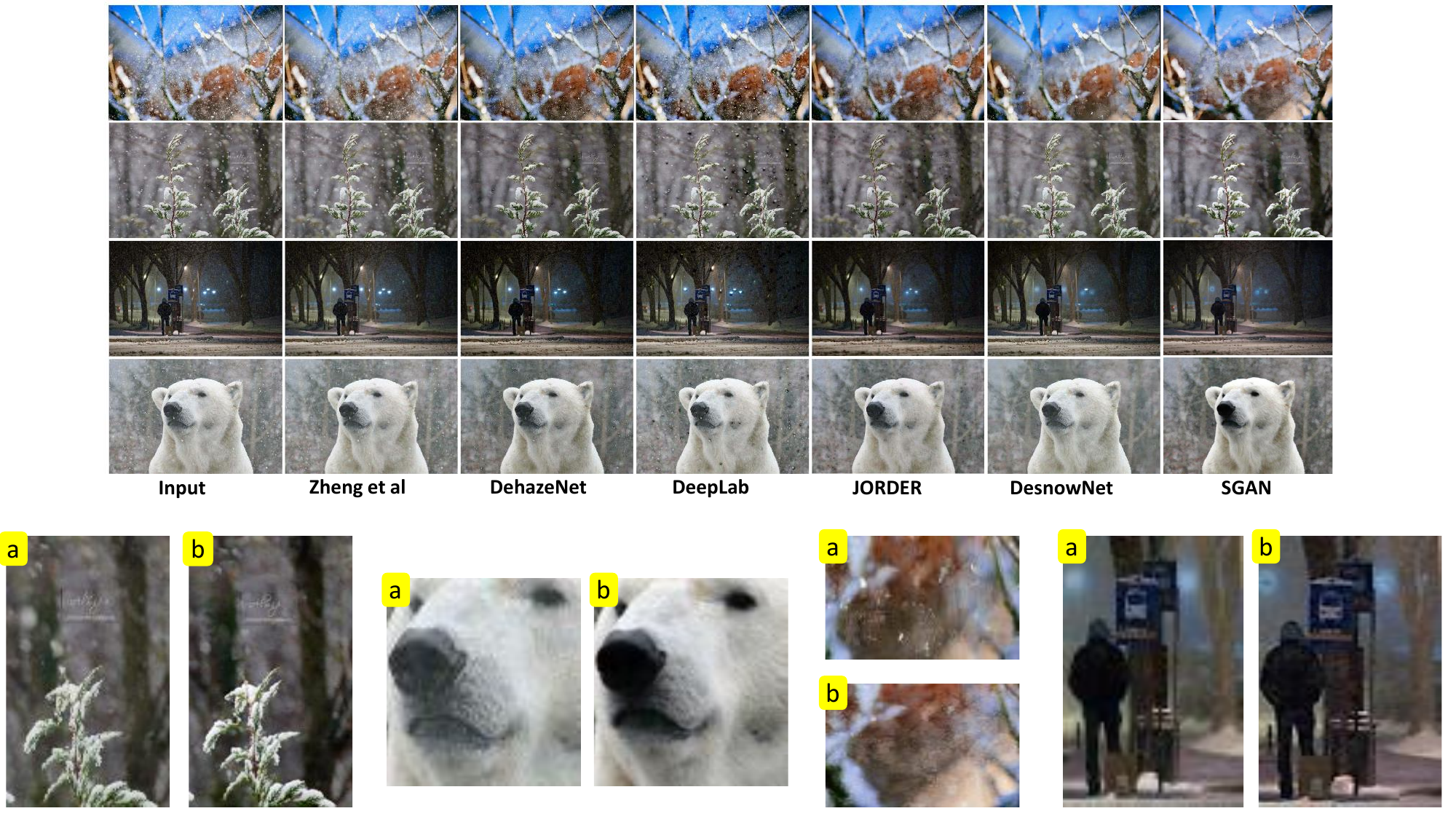}
    \caption{Visual results of the state-of-the-art desnowing approaches in addition to the proposed desnowing one SGAN where sub-images labelled by letter a are cropped from the results of DesnowNet while sub-images labelled by letter b are cropped from the results of our proposed approach.}
    \label{Fig3}
\end{figure*}

\begin{figure*}[!ht]
    \centering
    \includegraphics[width=17cm, height=4cm]{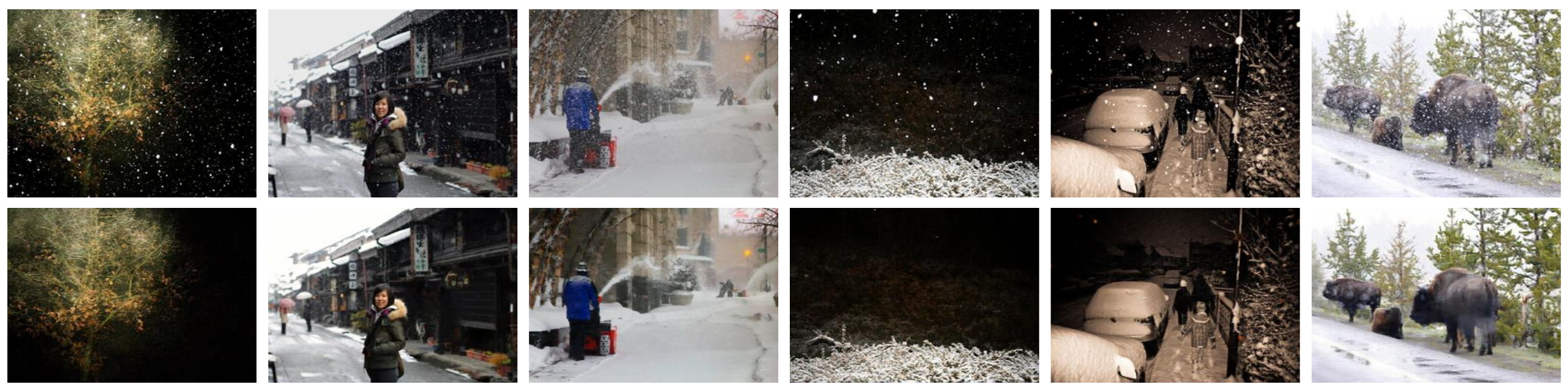}
    \caption{Examples of visual results generated by the proposed desnowing approach SGAN ($2^{nd}$ row).}
    \label{Fig4}
\end{figure*}

Nevertheless, our proposed SGAN outperforms the state-of-the-art approaches by average of 0.81 and 0.02 on PSNR and SSIM values, respectively. On the other hand, the images desnowed by SGAN are visually compared with the results presented in \citep{3}. As clearly shown in Fig. \ref{Fig3}, our proposed desnowing SGAN demonstrates better desnowing performance where it removes most snowflakes and preserves the high-frequency details of the input images. Furthermore, Fig. \ref{Fig4} shows more visual results extracted by the proposed SGAN, which demonstrates a very effective desnowing performance.  

\begin{table*}[htbp]
  \centering
  \caption{Performance comparison  of the proposed approach with other state-of-the-art approaches on snow100$K^2$}
    \begin{tabular}{p{6em} p{4.5em} p{4.5em} p{5em} p{5em} p{4.5em} p{4.5em} p{5em}}
    \toprule
    \multicolumn{1}{r}{} & \textit{Dehaze } & \textit{Deeplab} & \textit{JORDER} & \textit{DuRN-S-P} & \textit{DeSnow} & \textit{DS-GAN} & \textit{SGAN} \\
    \midrule
    Snow 100k-S & 24.96/0.88 & 25.94/0.87 & 25.62/0.88 & 32.27/0.94 & 32.33/0.95 & 33.43/0.96 & \textbf{33.74/0.96} \\

    Snow  100k-M & 24.16/0.86 & 24.36/0.85 & 24.97/0.87 & 30.92/0.93 & 30.86/0.94 & 31.87/0.95 & \textbf{32.59/0.97} \\

    Snow 100k-L & 22.61/0.79 & 21.29/0.77 & 23.41/0.80 & 27.21/0.88 & 27.16/0.89 & 28.06/0.92 & \textbf{29.48/0.95} \\

    \textit{Average} & 23.91/0.84 & 23.86/0.83 & 24.66/0.85 & 30.13/0.92 & 30.11/0.92 & 31.12/0.94 & \textbf{31.93/0.96} \\
    \bottomrule
    \end{tabular}%
  \label{Tab2}%
\end{table*}%

\begin{figure*}[!ht]
    \centering
    \includegraphics[width=17cm, height=8cm]{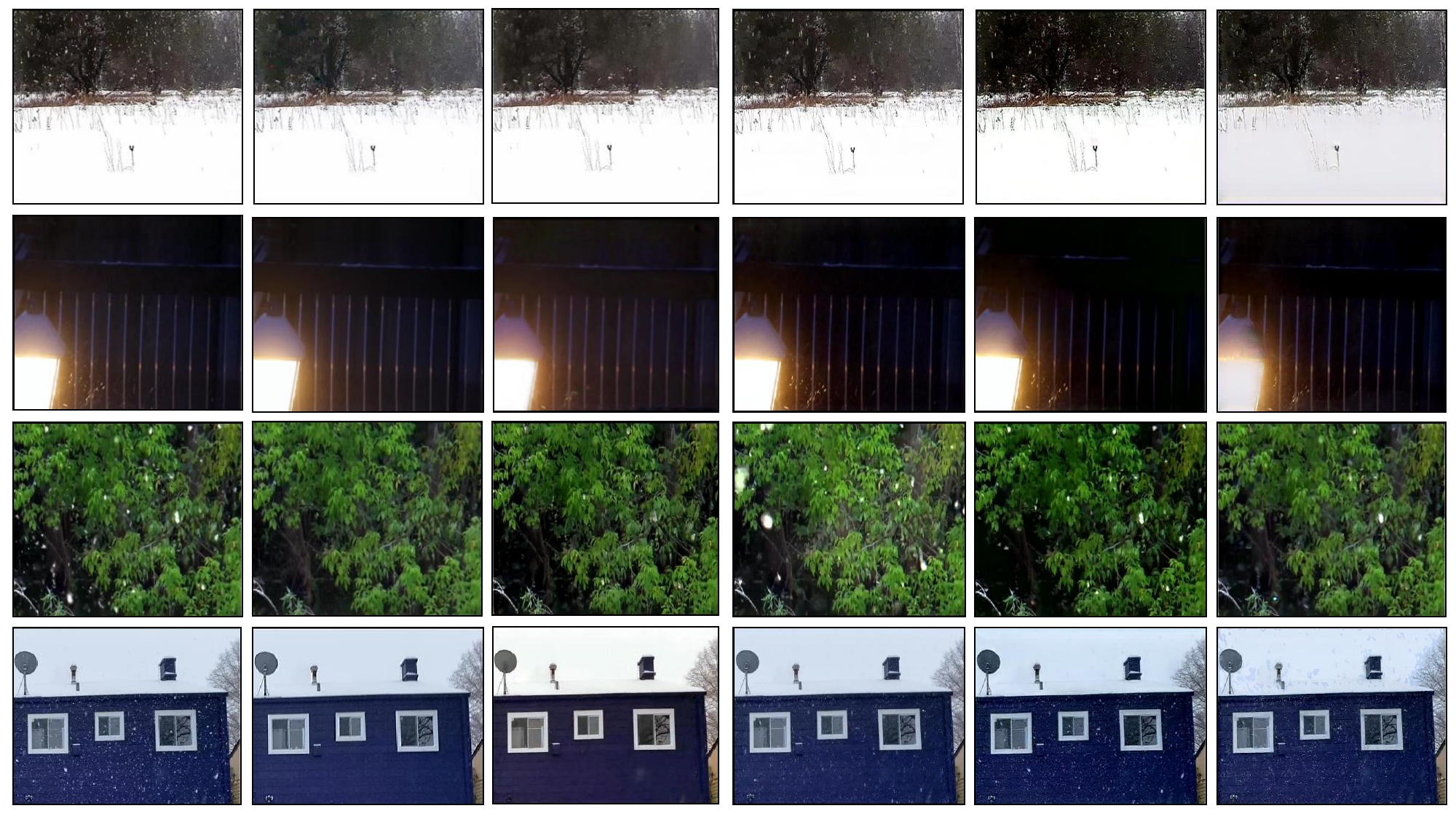}
    \caption{Visual comparison of the proposed desnowing SGAN several state-of-the-art desnowing approaches: form left to right: $1^{st}$ column: input images, $2^{nd}$ column: GTs, $3^{rd}$ column: SGAN, $4^{th}$ column: HDCWNet, $5^{th}$: TSK\&MAWR, $6^{th}$: JSTASR}
    \label{Fig5}
\end{figure*}

\textbf{Realistic dataset}: As mentioned in Section \ref{sect4}, the groundtruth images are estimated based on the low-rank approximation-based scheme and human supervision. The similarities between the desnowed images using SGAN and the groundtruth images are evaluated quantitatively and qualitatively. To validate the significance of the proposed dataset, we generate the desnowed images of five deep learning approaches using their original source codes: HDCWNet \cite{chen2021all}, TSK\&MAWR \cite{Chen2022MultiWeatherRemoval}, and JSTASR \cite{4}, in addition to 2 traditional GANs frameworks: U-net and Resnet. These approaches are trained on different datasets where HDCWNet and \\ TSK\&MAWR approaches are trained using CSD dataset \cite{chen2021all}, JSTASR \cite{4} is trained on SRRS dataset, while our proposed method, Resnet, and U-net are trained on Snow100k2. To evaluate the generalization capacity of the evaluated desnowing approaches, our proposed dataset is not used during training them. The PSNR and SSIM scores of the evaluated approaches are reported in Table \ref{sota-snow}. The PSNR and SSIM scores indicate that the proposed SGAN outperforms all evaluated approaches where it achieves a score of 35.56 and a score of 97.47 in terms of PSNR and SSIM respectively. Fig \ref{Fig5} illustrates several examples of the qualitative performance of all the tested approaches, including our proposed desnowing SGAN applied to realistic snowy images. It can be clearly seen that our proposed SGAN manged to remove most of snowflakes while significant number of snowflake residuals are still observed in the results estimated by existing state-of-the-art approaches.
\begin{table*}[htbp]
  \centering
  \caption{Performance comparison of the proposed approach with other state-of-the-art approaches on realistic snow dataset}
    \begin{tabular}{cccccccc}
    \toprule
          & Resnet  & U-net & HDCWNet & JSTASR & TSKMAWR (unified) & U-IncNet \\
    \midrule
    Realistic snow   & 26.85/0.9083 & 27.23/0.9170 & 22.87/0.6260 & 33.12/0.9681 & 32.25/0.8492 & \textbf{35.56}/\textbf{0.9747} \\
    \bottomrule
    \end{tabular}%
  \label{sota-snow}%
\end{table*}%
\begin{figure}[!ht]
  \centering
    \includegraphics[width=8cm, height=5.5cm]{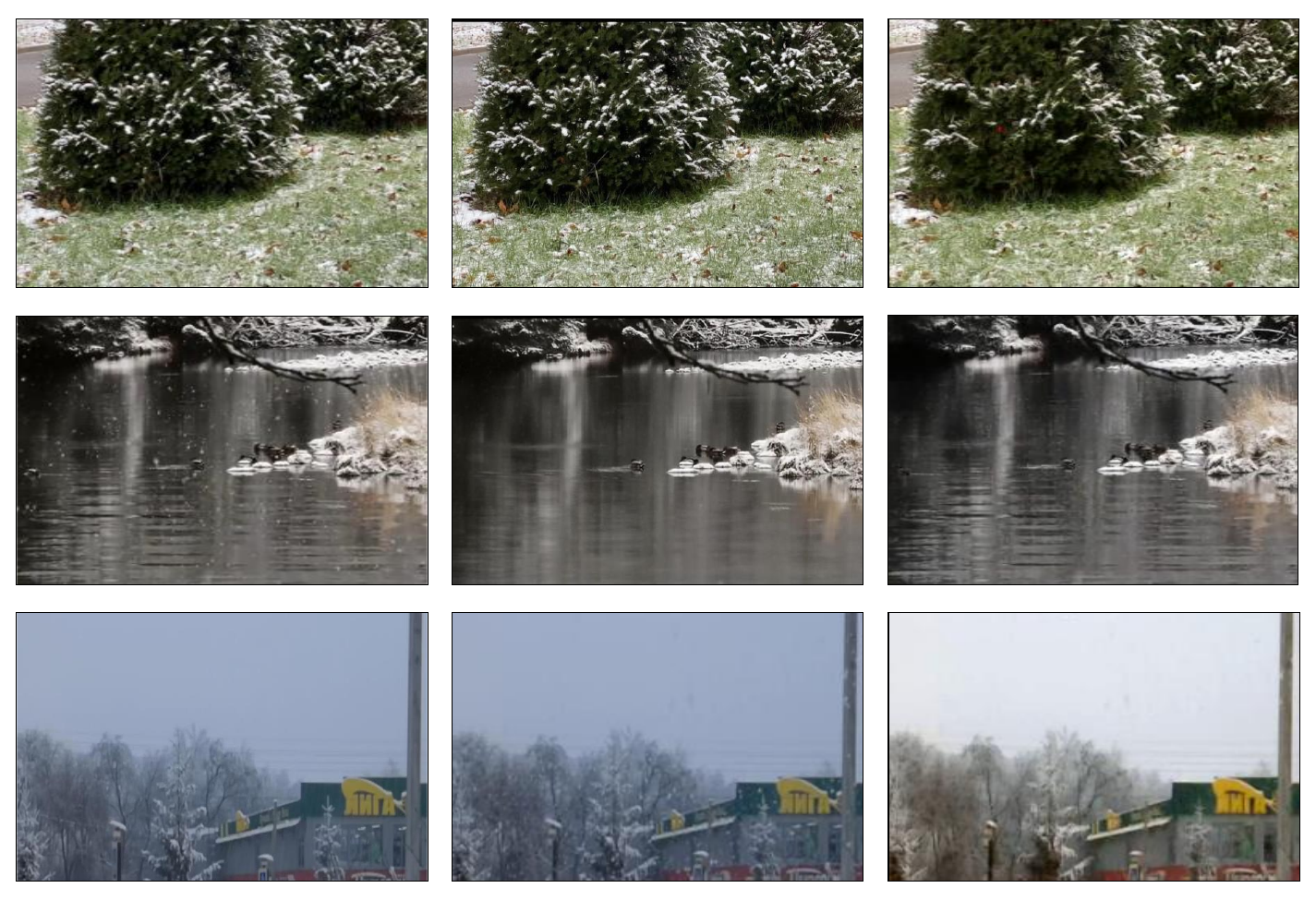}
    \caption{Visual examples of the differences between rainy images ($1^{st}$ column) processed by a learning-based approach ($2^{nd}$ column) and a spatiotemporal hand-crafted approach ($3^{rd}$ column).}
    \label{Fig6}
\end{figure}
However, several significant observations related to the visual results obtained by both video-based and image-based approaches have to be mentioned. As demonstrated in Fig. \ref{Fig6} ($1^{st}$ row), our groundtruth generator as a video-based approach can easily remove snowflakes based on their spatiotemporal characteristics while SGAN as an image-based approach lacks such an advantage leading to mistakenly removing some parts of the background that share the same appearance of snowflakes. On the other hand, video-based approaches fail to handle the veiling effect accompanying snow scenarios, as clearly shown in Fig 6 ($2^{nd}$ row). Furthermore, video based desnowing approaches suffer handling the challenge of dynamic background where they tend to oversmooth significant parts of background (see Fig. \ref{Fig6} ($3^{rd}$ row).
\subsection{Deraining experiment}

We conducted several experiments on both synthetic and realistic datasets to evaluate the performance of the proposed deraining approach and compare it with several state-of-the-art approaches.




\subsubsection{Ablation study}

We retrained the proposed RGAN on pairs of realistic rainy and rain-free images and conducted a comprehensive ablation study before testing it. 
\begin{table}[htbp]
  \centering
  \caption{Evaluation of the effectiveness of the proposed chained generators and the proposed loss }
    \begin{tabular}{ll}
    \toprule
    \multicolumn{1}{c}{\textbf{Derainer}} & \multicolumn{1}{c}{\textbf{score}} \\
    \midrule
    Resnet (11M \#Params) & 27.73/0.924 \\
    Unet   (54M \#Params)& 28.84/0.901 \\
    \textbf{U-IncNet}  (32M \#Params) & \textbf{30.04}/\textbf{0.948} \\
    \midrule
    \multicolumn{1}{c}{\textbf{Derainer+refinement}} & \multicolumn{1}{c}{\textbf{score}} \\
    \midrule
    U-IncNet+Unet & 31.36/0.958 \\
    U-IncNet+U-IncNet & 33.45/0.953 \\
    \textbf{U-IncNet+Resnet} & \textbf{35.14}/\textbf{0.958} \\
    \midrule
    \multicolumn{1}{c}{\textbf{Derainer+refinement+Loss function}} & \multicolumn{1}{c}{\textbf{score}} \\
    \midrule
    U-IncNet ($\mathbf{L}_\mathbf{1}^\mathbf{\nabla}$)+Resnet ($\mathbf{L}_{SSIM}$) & 36.24/0.979 \\
    \midrule
    \multicolumn{1}{c}{\textbf{Derainer+Loss function}} & \multicolumn{1}{c}{\textbf{score}} \\
    \midrule
    U-IncNet ($\mathbf{L}_\mathbf{1}^\mathbf{\nabla}$+$\mathbf{L}_{SSIM}$) & 32.24/0.949 \\
    \bottomrule
    \end{tabular}%
  \label{tab:addlabel}%
\end{table}%
To achieve this purpose, we have used the SPAnet dataset \cite{5}, which consists of 29,500 images extracted from 170 rainy videos that contain common urban scenes such as buildings, streets, and parks. The training images were divided into sub-images to increase the number of training images. A subset named Test1000 that contains 1000 realistic images with their groundtruths; is provided for testing.

Based on the fact that our deraining framework RGAN consists of two generators, the first ablation experiment concerns the selection of the best derainer among three GAN variants: ResNet, Unet, and U-IncNet. The best deraining results in terms of PSNR/SSIM are obtained by U-IncNet(30.04/0.94). In the second ablation study, the best derainer network U-IncNet is chained with three different variants of GAN and the estimated results are evaluated to select the best refinement network. Such results demonstrate that the best performance is achieved by the variant (U-IncNet, ResNet), where the resolution of the input image is well maintained by U-IncNet in addition to its efficiency in handling a variety of rain particle sizes thanks to the Inception.v3 module. Furthermore, the ResNet-based refinement generator $\mathbf{G}_{r}$ shows good performance in correcting the pixel saturation and the shift in the color that may appear in several images derained by U-IncNet based generator.The third experiment studies the impacts of the proposed losses $\mathbf{L}_\mathbf{1}^\mathbf{\nabla}$ and $\mathbf{L}_{SSIM}$ in which the variant (U-IncNet, Resnet) in the previous experiment was trained using the traditional GANs loss function. In this experiment, the derainer (U-IncNet) is trained using $\mathbf{L}_\mathbf{1}^\mathbf{\nabla}$ while the refiner (Resnet) is trained using $\mathbf{L}_{SSIM}$. 
\begin{figure}[!ht]
    \centering
    \includegraphics[width=8cm, height=6cm]{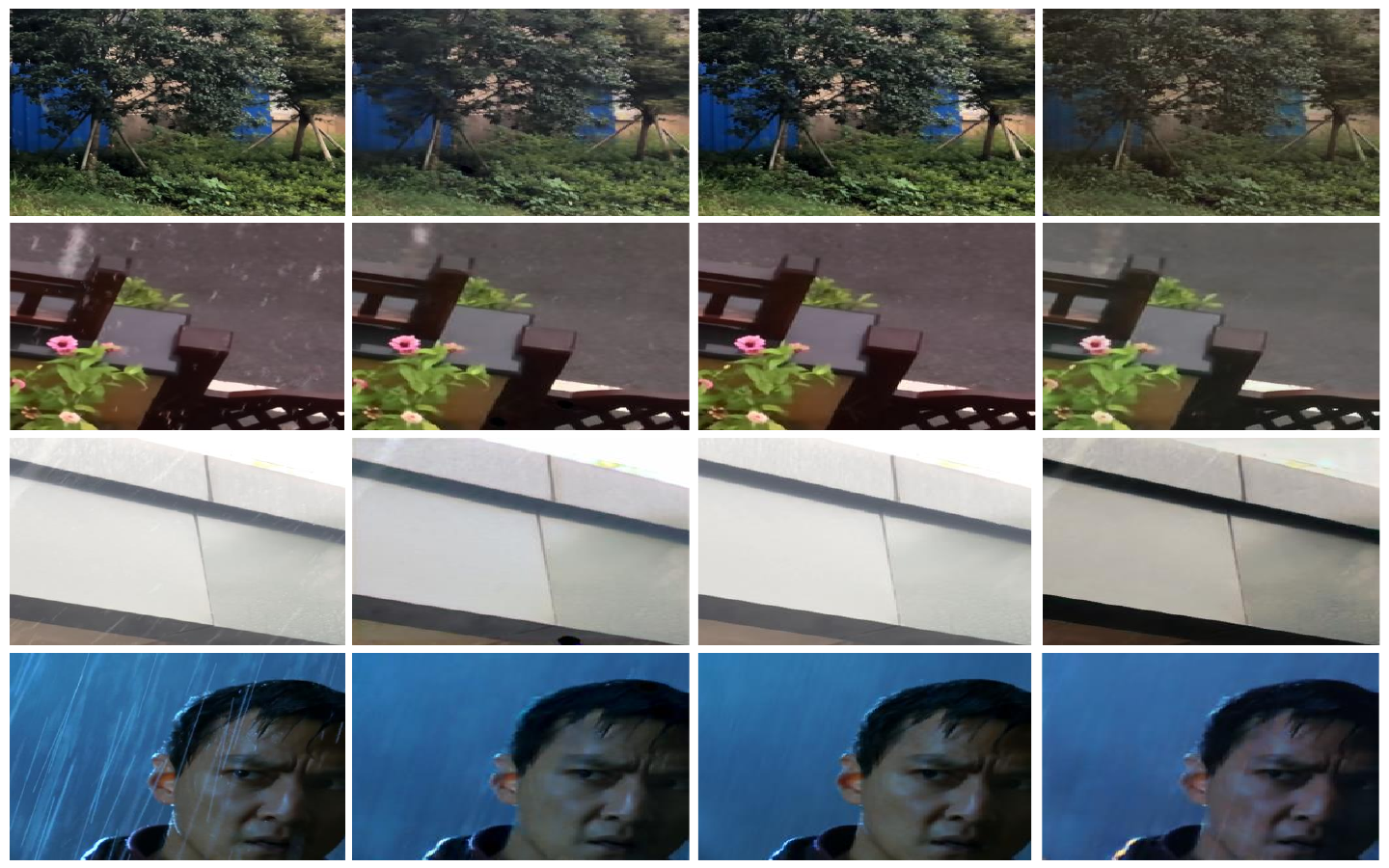}
    \caption{Visual examples of the differences between the results obtained by a generator that combines two loss functions and the images generated by an architecture of two chained generators: $1^{st}$ column shows input images, $2^{nd}$ column shows images after refinement, $3^{rd}$ column shows GTs, and $4^{th}$ column shows the results of the module (Derainer+Loss function)}
    \label{twovsone}
\end{figure}
The obtained result highlights the improvements brought about by the proposed loss $\mathbf{L}_\mathbf{1}^\mathbf{\nabla}$ where higher PSNR and SSIM scores are reached. The last experiment studies the impact of the architecture of chained generators with different losses versus a single generator with the same losses combined. The quantitative results as well as the qualitative ones (as shown in Fig. \ref{twovsone}) emphasizes the fact that chained generators offer a multi-stage enhancement which can not be provided by a single generator that combines both losses. These results also highlight the fact that despite the good deraining performance of the module (Derainer+Loss function), its results produce lower PSNR and SSIM socres compared to the scenario of two chained generators. This can be mainly contributed to the relatively low effectiveness in handling the color shift and the low saturation challenges.

After qualitatively and quantitatively comparing classic generators with the proposed variants, it is worth mentioning that the proposed collaboration architecture improves the performance of both generators compared to their individual performance in the case of a single generator-based network. Based on the facts concluded in this ablation study, the best variant (U-IncNet, ResNet) is used in the comparison study against several state-of-the-art approaches on both synthetic and realistic datasets.

\begin{table*}[htbp]
  \centering
  \scriptsize
  \caption{Performance comparison of RGAN with other state-of-the-art approaches on several synthetic subsets}
    \begin{tabular}{cccccccccc}
    \toprule
    dataset & RESCAN & SGCN  & PreNet & UMRL  & SPAnet & DCGAN & JRGR  & MPRNet & Proposed \\
    \midrule
    Rain 100 & 26.30/0.82 & 26.04/0.82 & 26.88/0.86 & 25.57/0.80 & 26.88/0.86 & 26.13/0.82 & 28.32/0.85 & 27.93/0.87 & \textbf{28.87}/\textbf{0.91} \\
    Test 100H & 23.63/0.92 & 24.93/0.89 & 24.55/69.85 & 24.78/0.68 & 24.50/0.69 & 23.26/0.74 & 27.62/0.81 & \textbf{27.67}/\textbf{0.92} & 26.67/0.84 \\
    Test 1200 & 26.52/0.88 & 30.54/0.92 & 31.91/0.95 & 26.38/0.88 & 31.87/0.95 & 25.60/0.87 & 29.26/0.91 & \textbf{33.12}/0.94 & 32.84/\textbf{0.97} \\
    Test 2800 & 27.47/0.90 & 27.51/0.91 & 31.10/0.95 & 26.52/0.89 & 31.13/0.95 & 27.38/0.91 & 28.53/0.92 & 30.99/0.95 & \textbf{33.41}/\textbf{0.96} \\
    Average & 25.98/0.84 & 27.26/0.89 & 28.61/0.87 & 25.78/0.82 & 28.61/0.87 & 25.77/0.84 & 28.43/0.88 & 29.92/0.92 & \textbf{30.44}/\textbf{0.92} \\
    \bottomrule
    \end{tabular}%
  \label{synthetic}%
\end{table*}%

\begin{table*}[htbp]
  \centering
  \scriptsize
  \caption{Performance comparison of RGAN with other state-of-the-art approaches on realistic TEST1000 subset}
    \begin{tabular}{cccccccclc}
    \toprule
    dataset & RESCAN & SGCN  & PreNet & UMRL  & SPAnet & DCGAN & JRGR  & \multicolumn{1}{c}{MPRNet} & Proposed \\
    \midrule
    Test 1000 & 31.30/0.93 & 33.77/0.95 & 32.32/0.94 & 24.04/0.85 & 32.32/0.94 & 27.74/0.90 & 23.65/0.81 & 33.81/0.95 & \textbf{34.21}/\textbf{0.96} \\
    \bottomrule
    \end{tabular}%
  \label{Tab5}%
\end{table*}%

\begin{table*}[htbp]
  \centering
  \scriptsize
  \caption{Average NIQE and BRISQUE results of RGAN in comparison with the state-of-the-art approaches}
    \begin{tabular}{cccccccccc}
    \toprule
    dataset & RESCAN & SGCN  & PreNet & UMRL  & SPAnet & DCGAN & JRGR  & MPRNet & Proposed \\
    \midrule
    RID   & 5.08/32.21 & 6.35/28.27 & 6.53/30.23 & 5.42/24.12 & 6.53/3.22 & 8.9/25.56 & 7.81/25.52 & 6.37/29.04 & \textbf{4.84}/\textbf{23.71} \\
    RIS   & 5.30/29.14 & 5.36/30.77 & 5.54/30.96 & 5.52/27.22 & 5.55/30.97 & 5.74/31.46 & 5.26/29.90 & 5.59/31.03 & \textbf{5.06}/\textbf{26.27} \\
    \bottomrule
    \end{tabular}%
  \label{Tab6}%
\end{table*}%

\subsubsection{Comparison with the state-of-the-art approaches}
To perform this comparison, we carry on the same training protocol followed in \citep{35}, where a training set of 10500 images that are collected from four synthetic datasets; is employed. For better generalization ability, we have added 3000 images from SPAnet dataset that contain pairs of realistic rainy images and groundtruths. For fair comparison, we retrained the state-of-the-art deraining models (if the original training source code is available) and tested on the following synthetic subsets: \textbf{Test100} \citep{19}, \textbf{TestH} \citep{19}, \textbf{Test2800} \citep{20}, and \textbf{Test1200} \citep{36} in addition to the realistic subset: TEST1000 from SPAnet dataset. The evaluated approaches include: RESCAN \citep{37}, successive graph-CNN (SGCN) \citep{fu2021successive}, PreNet \citep{40}, UMRL \citep{38}, SPAnet \cite{5},  DerainCycleGAN (DCGAN) \citep{42}, JRGR \cite{ye2021closing},  and MPRNet \cite{Zamir2021MPRNet}.


\begin{figure*}[!ht]
    \centering
    \includegraphics[width=14cm, height=21cm]{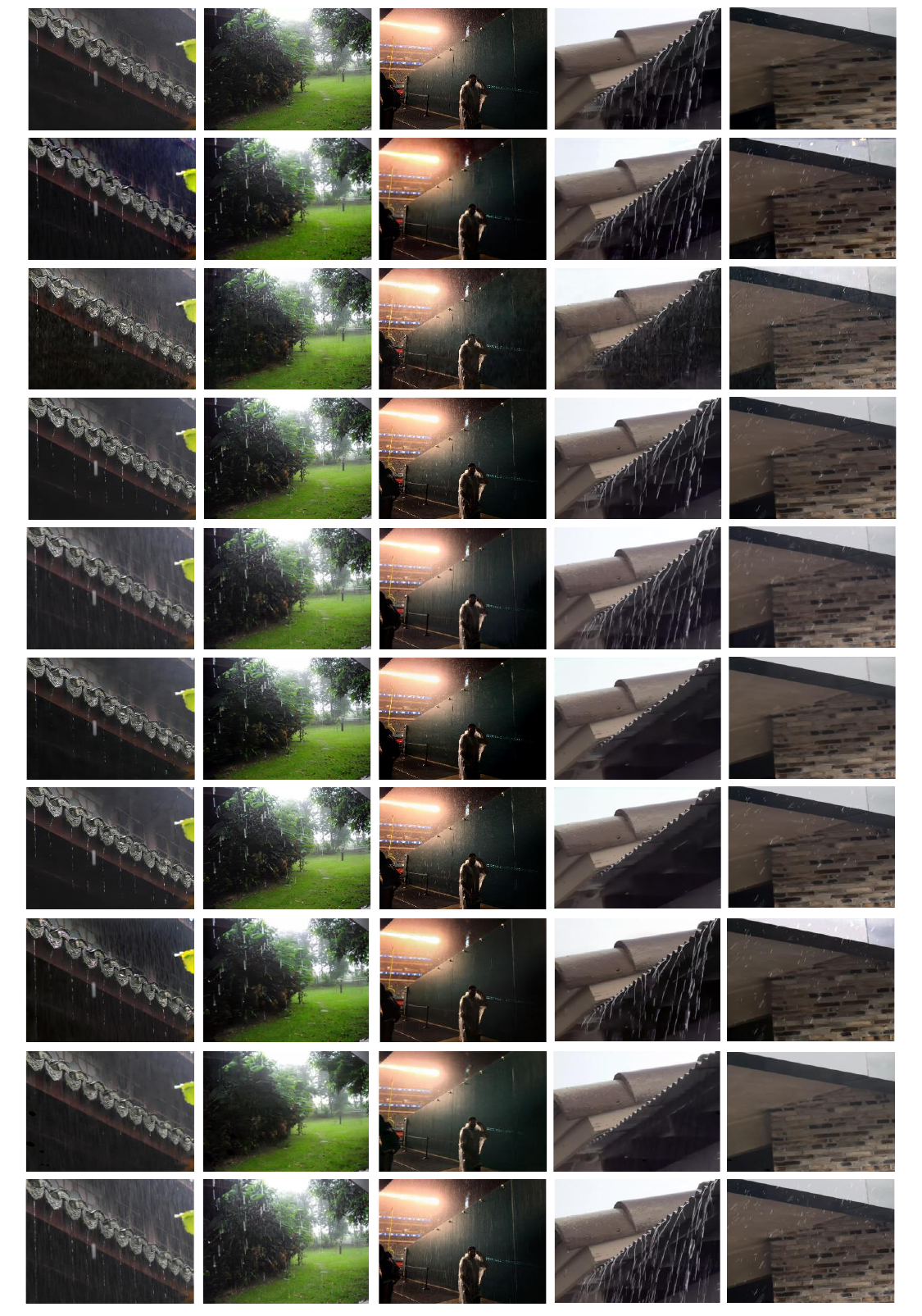}
    \caption{Visual comparison of the proposed deraining RGAN with the state-of-the-art approaches on realistic rainy images from internet and TEST1000 subset.From top to bottom: RESCAN, PreNet, UMRL, DCGAN, JRGR, SGCN, SPAnet, MPRNet, RGAN, and rainy input}
    \label{Fig7}
\end{figure*}

\textbf{Synthetic dataset}: As mentioned earlier, learning methods trained on synthetic datasets are often impractical due to the complexity of the appearance of the rain patterns in many real-world scenarios. However, testing a deraining approach on synthetic images helps evaluating the efficiency of such deraining approach. The calculated PSNR/ SSIM values of the evaluated approaches including the proposed one are reported in Table \ref{synthetic}. The results show that the proposed deraining approach achieves remarkable improvements over the state-of-the-art approaches in terms of PSNR and SSIM values. 

\textbf{Realistic dataset}: The results obtained by the best variant (U-IncNet, ResNet) are compared with the state-of-the-art techniques (see Table \ref{Tab5}) on TEST1000 realistic subset. The proposed RGAN outperforms the best reported technique by 1.54 and 0.012 in terms of PSNR and SSIM values respectively. Moreover, the visual results obtained when testing several deraining approaches including the proposed one on Test1000 and real internet subsets are shown in Fig \ref{Fig7}.
 The visual results show the efficiency of the proposed RGAN in deraining the rainy images while preserving the color and the high-frequency details. It is worth noting that despite their good deraining performances, several state-of-the-art approaches produce lower PSNR and SSIM values due to minor changes in the color distributions. This sheds lights on the need of a new evaluation metric that can evaluate the deraining performance of a certain approach independently from pixel similarity-based evaluation. 
The last experiment concerns the evaluation of the proposed RGAN based on non-reference metric in the case of realistic rain datasets that have no groundtruths. For this experiment, two non-reference metrics are used: Naturalness Image Quality Evaluator (NIQE), and Blind/Referenceless Image Spatial Quality Evaluator (BRISQUE) where smaller values indicate good restoration performance. Two outdoor datasets which capture several driving scenes in rainy weather are used: Rain in Driving (RID) and Rain in Surveillance (RIS) datasets. The results listed in Table \ref{Tab6} show that the proposed RGAN achieves the best average NIQE and SSEQ values on both datasets by a significant margin.



\section{Conclusion}

In this paper, a global framework that handles two main types of image degradation caused by snow and rain particles; is proposed. 
In the desnowing procedure, a single generator-based GAN with a novel architecture that takes full advantage of both ResNet and U-net architectures is introduced. On the other hand, the deraining approach consists of a collaboration structure between two generators that are trained to remove the rain strikes/drops in addition to a refinement stage for better visual results. Both the desnowing and deraining GANs are spatially guided by unique cost functions that respect the unique characteristics of each degradation type. Both approaches are trained and tested by means of synthetic and realistic datasets. Due to the absence of a realistic snow dataset in the literature, we presented a low-rank approximation based hand-crafted technique that helps producing clean images extracted from a sequence of snowy images. Based on this technique, a small subset that contains pairs of snowy/clean images is provided to evaluate the proposed desnowing approaches quantitively and qualitatively. The conducted experiments show that both proposed desnowing and deraining approaches perform favorably against the state-of-the-art approaches on both synthetic and realistic test subsets. \\ \\

\bibliographystyle{apalike}
\bibliography{references}   

\end{document}